\documentclass[acmsmall, nonacm=True, screen]{acmart}
\AtBeginDocument{%
  \providecommand\BibTeX{{%
    \normalfont B\kern-0.5em{\scshape i\kern-0.25em b}\kern-0.8em\TeX}}}
    
\usepackage[linesnumbered,ruled,vlined]{algorithm2e}
\usepackage{amsmath}
\newcommand{\pluseq}{\mathrel{+}=}

\begin{document}

\title{Reward-Based 1-bit Compressed Federated Distillation on Blockchain}

\author{Leon Witt}
\affiliation{%
  \institution{Tsinghua University}
  \city{Beijing}
  \country{China}}
\affiliation{%
  \institution{Fraunhofer Heinrich Hertz Institute}
  \city{Berlin}
  \country{Germany}}
\email{leonmaximilianwitt@gmail.com}

\author{Usama Zafar}
\affiliation{%
  \institution{Tsinghua University}
  \city{Beijing}
  \country{China}}
\email{usamazf@outlook.com}

\author{KuoYeh Shen}
\affiliation{%
  \institution{Tsinghua University}
  \city{Beijing}
  \country{China}}
\email{kuoyehs@gmail.com}

\author{Felix Sattler}
\affiliation{%
  \institution{Fraunhofer Heinrich Hertz Institute}
  \city{Berlin}
  \country{Germany}}
\email{felix.sattler@hhi.fraunhofer.de}

\author{Dan Li}
\affiliation{%
  \institution{Tsinghua University}
  \city{Beijing}
  \country{China}}
\email{tolidan@tsinghua.edu.cn}

\author{Wojciech Samek}
\affiliation{%
  \institution{Fraunhofer Heinrich Hertz Institute}
  \city{Berlin}
  \country{Germany}}
\email{wojciech.samek@hhi.fraunhofer.de}

\renewcommand{\shortauthors}{Witt et al. - Reward-Based 1-bit Compressed Federated Distillation on Blockchain}

\begin{abstract}
The recent advent of various forms of Federated Knowledge Distillation (FD) paves the way for a new generation of robust and communication-efficient Federated Learning (FL), where mere soft-labels are aggregated, rather than whole gradients of Deep Neural Networks (DNN) as done in previous FL schemes. This security-per-design approach in combination with increasingly performant Internet of Things (IoT) and mobile devices opens up a new realm of possibilities to utilize private data from industries as well as from individuals as input for artificial intelligence model training. Yet in previous FL systems, lack of trust due to the imbalance of power between workers and a central authority, the assumption of altruistic worker participation and the inability to correctly measure and compare contributions of workers hinder this technology from scaling beyond small groups of already entrusted entities towards mass adoption. This work aims to mitigate the aforementioned issues by introducing a novel decentralized federated learning framework where heavily compressed 1-bit soft-labels, resembling 1-hot label predictions, are aggregated on a smart contract. In a context where workers' contributions are now easily comparable, we modify the Peer Truth Serum for Crowdsourcing mechanism (PTSC) for FD to reward honest participation based on peer consistency in an incentive compatible fashion. Due to heavy reductions of both computational complexity and storage, our framework is a fully on-blockchain FL system that is feasible on simple smart contracts and therefore blockchain agnostic. We experimentally test our new framework and validate its theoretical properties.   
\end{abstract}



\maketitle

\section{Introduction}
The increasing demand for confidential \textit{Machine Learning} (ML) led to the advent of \textit{Federated Learning} (FL), where complex models such as \textit{Deep Neural Networks} (DNNs) are trained in parallel on end devices while data remains local at all times. \textit{Federated Averaging} (\texttt{FedAvg}) \cite{BrendanMcMahan2017} is a widely applied algorithm for FL, where locally trained models $\theta_i$ get aggregated on a central location to form a global model. 
Even though first use-cases \cite{Ramaswamy2019, Hard2018, Yang2019} hint at the potential of utilizing untapped raw data and computational power, communication overhead due to the gradient size of modern DNNs remains - among other issues \cite{Li2019} - a major bottleneck of \texttt{FedAvg}.
Beyond technical issues, the lack of trust due to the imbalance of authority between workers and the central server as well as the lack of a practical reward system for contributions of a worker hinder this technology from scaling beyond small groups of already entrusted entities towards mass adoption. As we will demonstrate in this paper, these limitations can be overcome.

\textbf{Blockchain to ensure equal power.} General purpose blockchain systems \cite{wood2014ethereum,polkadotwhitepaper, HyperledgerFabric} have the potential to mitigate the first issue by ensuring trust through their inherent properties of immutability and transparency of a distributed ledger, thereby enabling decentralized federations to mitigate dependencies on a central authority \cite{nature2021,DeepChain, GFL, Biscotti, Blockchained}.

 \textbf{Mechanism Design to incentivize participation.}     In order to incentivize participation, workers have to be rewarded for their contributions. An appropriately designed mechanism ensures a desired equilibrium when every worker acts rationally and in its own best interest. Such a mechanism has low complexity and is self-organizing, avoiding the need for Trusted Execution Environments \cite{TXX} or cryptographic schemes. In a FL context, this demands a carefully designed reward strategy based on the quality of contributions. Yet, comparing and evaluating workers gradient-updates of \texttt{FedAvg} remains challenging \cite{9367484}. So far, few holistic decentralized gradient-aggregation based FL system with reward mechanisms have been introduced \cite{DeepChain, Liu2020, Biscotti, FLChain}. 

\textbf{Federated Distillation to reduce communication overhead.}
As it is costly to store large amounts of data or perform complex computations on Blockchain due to its shared ledger architecture in which every node has to replicate every computation and all stored information, the data intensive gradient aggregation process of \texttt{FedAvg} typically can not be embedded in simple and lightweight Smart Contract Architectures \textit{on-chain}. The aggregation process therefore either requires novel \textit{application specific blockchain systems} (ASBS) \cite{DeepChain, Biscotti, FLChain, Liu2020} or off-chain aggregation and evaluation \cite{FLBlockchainMAIN}. Both paradigms cause drawbacks: ASBS for FL cause huge overhead in terms of storage and complexity and therefore restrict practicality. Paradigms where the data is being aggregated off-chain face a potential data availability problem, lack of a decentralized reward mechanism or a single point of failure. 
Recently proposed Federated Knowledge Distillation (FD) frameworks \cite{Lin2020,Itahara2020,Seo2020} introduce an alternative paradigm to gradient-aggregation schemes like \texttt{FedAvg}. The FD process aggregates soft-label predictions on a public unlabelled dataset $X^{p u b}$ with communication proportional to $\mathcal{O}\left(|X^{p u b}| \operatorname{dim}{\mathcal{C}}\right)$ instead of gradients $\mathcal{O}\left(|\theta|\right)$ in \texttt{FedAvg}, where $|X^{p u b}|$ is the size of the public unlabelled dataset and $\operatorname{dim}{\mathcal{C}}$ is the dimension of the soft-labels, e.g. the number of different classes that are predicted. The amount of information necessary to be exchanged can be orders of magnitude lower in FD compared to \texttt{FedAvg} and
recent works of \cite{Sattler2020} suggests that the soft-labels of a classification task within the FD process can be further compressed to 1bits without sacrificing top-1 accuracy, as long as $|X^{p u b}|$ is sufficiently large. This not only reduces the default float32 soft-labels by 32-fold, but 1-bit quantized soft-labels resemble a 1-hot prediction for a specific task, which therefore allows for efficient encoding schemes (e.g. integer encoding) necessary for blockchain architectures where floating point numbers are not supported \cite{wood2014ethereum}. Most importantly, contributions by workers become deterministic and easily comparable. \\

This paper introduces a novel reward-based Federated Learning paradigm based on Federated Knowledge Distillation of 1-bit quantized soft-labels. With the aformentioned reductions of storage and complexity by orders of magnitude, our framework is able to aggregate and reward participation within a blockchain agnostic smart contract without the need for ASBS. Capitalizing on the comparability of contributions, we further introduce a peer consistency based mechanism called \textit{Peer Truth Serum for Federated Distillation (PTSFD)}, extending the Peer Truth Serum for Crowdsourcing (PTSC) \cite{PTSC} for FD context. PTSFD ensures a strategy profile in which all workers exert high effort and report their results truthfully as the most profitable equilibrium, paving the way for potential mass adoption of confidential and robust FL. We theoretically validate the presence of an incentive compatible equilibrium and perform a systematic experimental evaluation, demonstrating that (i) under different FL environments participation in the federation leads to a substantial increase in local model accuracy for every worker (even if the data is non-iid), (ii)  the reward distribution is positively correlated with exerted effort and (iii) the framework is robust against collusion and malicious behavior.


\section{Background and related work}
\subsection{Federated Averaging}

The classical algorithmic approach to Federated Learning problems is Federated Averaging. In Federated Averaging the training is conducted in multiple communication rounds following a three step protocol:
\begin{enumerate}
    \item At the beginning of each round, the central server selects a subset of the client population and broadcasts a common model initialization $\theta$ to them.
    \item Starting from the common initialization, the selected clients individually perform iterations of stochastic gradient descent over their local data to improve their local models, resulting in an updated model $\theta_{i}$ on every client.
    \item The updated models are then communicated back to the server, where they are aggregated (e.g. by an averaging operation) to create a global model, which is used as initialization point for the next communication round.
\end{enumerate}

Every communication round of Federated Averaging thus involves the upstream and downstream communication of a complete parametrization of the jointly trained model $\theta$ between all participating clients. In many practical applications these neural network parametrizations may contain multiple millions to billions of individual parameters. For instance, the widely popular ResNet-5 8 contains over 23 million parameters. For natural language processing tasks even larger models are used, with the carefullys GPT-3 9 clocking in at .5 billion parameters. Generally, both theoretical \cite{kidger2020uat, chong2020uat} and empirical \cite{huang2019gpipe} evidence suggests that the performance of neural network models correlates positively with their size. Although a wide variety of methods to reduce the model size in Federated Averaging have been proposed like neural network pruning \cite{lecun1990pruning} , and other lossy \cite{courbariaux2015compress, li2016compress, konevcny2016lowrank, sattler2018sparse, xu2020compress, SatTNNLS20b}, and loss-less compression techniques \cite{neumann2020deepcabac, wiedemann2020deepcabac}, the fundamental issue of scaling to larger models persists in prohibiting the use of Distributed Ledger Technologies for storing or aggregating models.      

\begin{figure}[th]
    \centering
    \includegraphics[width=\textwidth]{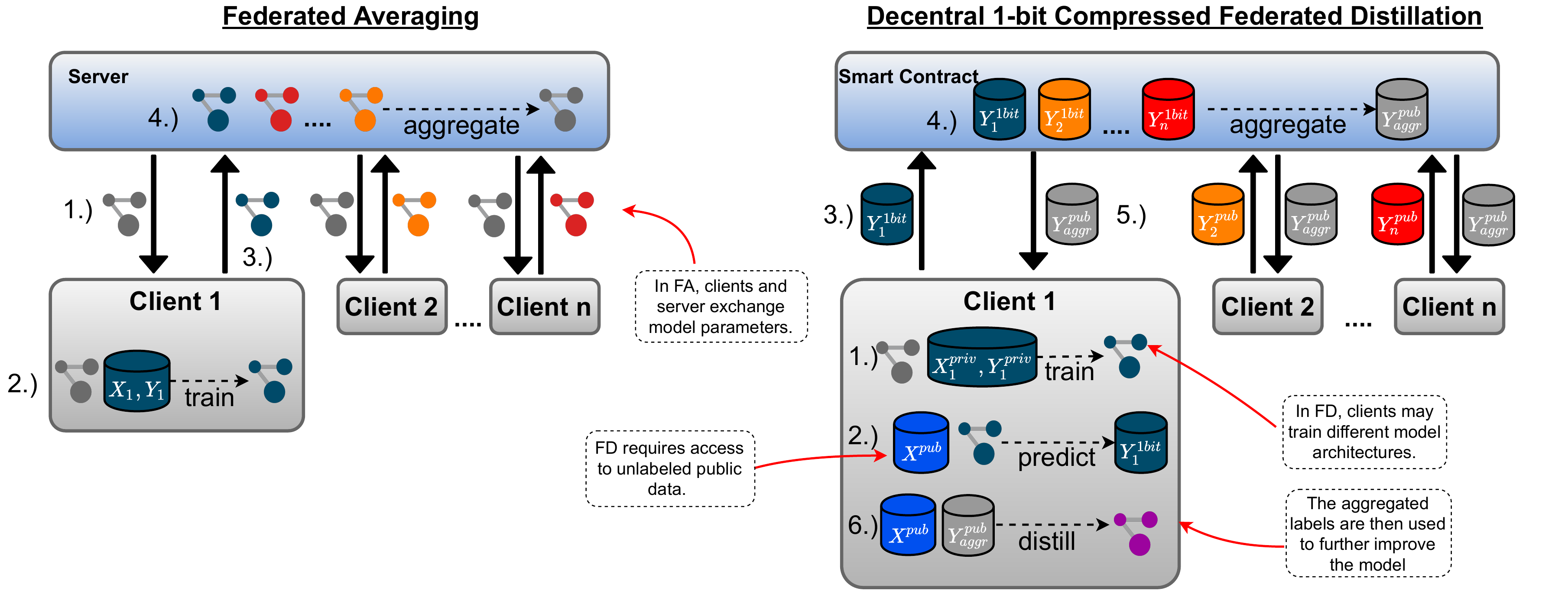}
    \caption{Federated Learning vs. Decentral 1-bit Compressed Federated Distillation on Blockchain.}
    \label{fig:FLvsFD}
\end{figure}

\subsection{Federated Distillation}

\label{Sec:Federated Distillation}
 Federated Distillation \cite{jeong2018distill, Lin2020, Itahara2020} is a recently developed FL algorithm implementing a very different knowledge exchange strategy. Here, soft-label predictions $Y_i^{pub}$ obtained by applying the updated local model to a public distillation dataset $X^{pub}$ are the carrier of information about the local model improvement ($\theta_i \to \theta_i + \Delta\theta_i$), i.e.,
\begin{equation}
Y_i^{pub} = \{f_{\theta_i + \Delta\theta_i}(x)\ |\ x \in X^{pub}\}.    
\end{equation}

Since this type of knowledge communication scales with the size of the distillation dataset and not with the number of model parameters, it can lead to a significant reduction in communication overhead \cite{Sattler2020}, especially for very large models. Furthermore, Federated Distillation allows clients with different model architectures to participate in the joint training process.

Different variants of Federated Distillation have been proposed in the scientific literature. In this work, we will modify a recently proposed, highly communication-efficient FD method \cite{Sattler2020}, termed {\it Compressed Federated Distillation (CFD)}, which is based on the multi-round protocol developed in \cite{Itahara2020,jeong2018distill}. In our modified version of CFD, every client performs the following steps in each communication round:
\begin{enumerate}
\item {\bf Train} on local datasets and improve  model $\theta_i = \theta + \Delta\theta_i$ by using $X_{i}^{priv}, Y_{i}^{priv}$.
\item {\bf Predict labels} $Y_i^{1bit}$ by using the improved model $\theta_i$ on $X^{pub}$ to compute soft-labels ${Y}_i^{pub}$ and perform 1bit quantization ${Y}_i^{1bit} = Q_{1bit}(Y^{pub}_i)$.
\item {\bf Upload} the integer encoded compressed soft-labels to the smart contract (in a two step commit-reveal fashion outlined in Algorithm ~\ref{Alg:Commit Reveal}).
\item {\bf (Blockchain) Aggregate} predictions $Y_{aggr}^{pub}$ by majority vote over all ${Y}_i^{pub}$.
\item {\bf Download} the aggregated predictions $Y_{aggr}^{pub}$ from the blockchain.
\item {\bf Distill} the current model $\theta$ by using $X^{pub}$ and  $Y_{aggr}^{pub}$.
\end{enumerate}

The authors of \cite{Sattler2020} showed that CFD largely reduces the information necessary for exchange by quantization $Q$ and the use of small public distillation dataset (e.g., random subset selection). The savings are in the order of two orders of magnitude when compared to Federated Distillation, and more than four orders of magnitude when compared to parameter averaging based techniques like Federated Averaging.  The possibility to apply binary soft-label quantization, i.e., $Q_b$ with $b=1$, ensure three important properties for a decentralized CFD on Blockchain, namely
\begin{itemize}
    \item It reduces the amount of information processed in the aggregation process heavily.
    \item It makes contributions by workers explicit and comparable. 
    \item It supersedes the need for additional encryption like noise inducing Differential Privacy or computational heavy secure multiparty computation.
\end{itemize}

\subsection{Blockchain Technology in FL context}
Blockchain was initially introduced with Bitcoin by Satoshi Nakamoto in 2008 \cite{nakamoto2008bitcoin}. Blockchain is referred to as a distributed ledger managed by nodes in a peer to peer network, where cryptographic links of information ensure resistance to modification and immutability. The network is governed by a consensus mechanism \cite{8693657} among peers which supersedes the need for central coordination. The advent of general purpose blockchains \cite{wood2014ethereum} with smart contract functionality supporting Turing-completeness allow for a decentralized, immutable and transparent business logic atop of blockchain. This technology is able to mitigate open problems of FL environments due to its inherent properties, namely: \begin{enumerate}
    \item \textbf{Decentralization.} In server-worker architectures, workers are exposed to a power imbalance and single point of failure. A malicious server could (i) exclude workers arbitrarily or (ii) withhold reward payments. Furthermore, a server-worker design is not suitable for an environment where multiple entities share a common and equal interest in advancing their respective models. The decentral property of blockchain systems ensures a federal systems for entities with equal power without the need for a central server.  
    \item \textbf{Transparency and Immutability.} Since every peer in the system shares the same data, data on blockchain can only be updated and never deleted. A transparent and immutable reward logic in an FL context ensures trust on the worker side. On the other hand, each worker is audited and can therefore be held accountable for malicious behavior.
    \item \textbf{Cryptocurrency.} Many general-purpose blockchain systems come with cryptocurrency functionality, e.g. the option to implement payment schemes within the business logic of the smart contract. Based on a reward mechanism of the FL system, workers can be rewarded immediately, automatically and deterministically. 
\end{enumerate}

To analyze Blockchain systems, we categorize into \textit{Application Specific Blockchain Systems (ASBS)} and \textit{General Purpose Blockchain Systems (GPBS)}. Both systems can be either permissioned or public. Blockchains which have to be adopted to a specific use-case require a novel infrastructure which causes overhead in terms of complexity at the development, deployment and operations level of such a system. GPBS are limited due to restricted virtual machines and predefined consensus layers but allow for easy development, deployment and operation utilizing already existing frame\-works, e.g.,  \cite{buterin2013whitepaper,polkadotwhitepaper,avaxwhitepaper,cosmoswhitepaper, HyperledgerFabric}. Due to large gradient sizes, complex comparability measures of contributions as well as cryptographic methods to ensure confidentiality of \texttt{FedAvg}, holistic blockchain based decentral FL systems require ASBS \cite{DeepChain, Biscotti, FLChain}. On the contrary, our decentral FL system is based on 1bit compressed FD instead of \texttt{FedAvg}, which reduces storage and computational complexity by orders of magnitude in comparison, makes contributions easily comparable and mitigates vulnerability to model inversion attacks and therefore may supersede the need for additional cryptography. These properties allow our holistic, completely on-blockchain framework to be blockchain agnostic and reside atop of even heavily restricted GBPS like Ethereum \cite{wood2014ethereum}. Even though theoretically possible, many promising public blockchain projects are still in their technological infancy, e.g. either are yet to implement smart contract functionality (e.g. Cardano \cite{Ouroboros}, IOTA \cite{iotawhitepaper}) or face scalablility restrictions (e.g. Ethereum), which makes it economically infeasible and causing scalability issues to deploy our system atop of public blockchains as of now, for the high cost of transaction fee and the limitation of transaction per second. Technical advancements on Ethereum like sharding\footnote{Ethereum Sharding endevours https://ethereum.org/en/eth2/shard-chains/} or layer 2 solutions for off-chain computation like optimistic rollups\footnote{https://docs.ethhub.io/ethereum-roadmap/layer-2-scaling/optimistic\_rollups/} and zero-knowledge rollups\footnote{https://docs.ethhub.io/ethereum-roadmap/layer-2-scaling/zk-rollups/} may change that in the short-term future.

\subsection{Related Work}
Our proposed FL framework combines a (i) Reward Mechanism and (ii) Decentralization. The task to classify $X^{p u b}$ resembles a (iii) Crowdsourcing Contest. 

\textbf{Crowdsourcing Contests.}
A crowdsourcing contest describes a game-theoretic framework where workers invest irreversible and costly efforts towards winning a reward from the requester, which is allocated based on relative performance \cite{ContestTheory}. In a crowdsourcing environment, workers are recruited anonymously through the Internet, so a major issue is how to ensure that their answers are accurate. Recent works on Peer Prediction \cite{CorrelatedAgreement, PeerPredictionHeterogenous, PTSC} have been studied to elicit truthful information from agents without any objective ground truth against which to score reports. In addition, many works have analyzed the optimal strategies between maximizing requesters utility and workers profit \cite{Optimaldesign, heterogeneouscrowdsourcing, endogenous, payment,tullock, crowdsourcingcontests}. The authors of \cite{Optimaldesign} showed that in an all-pay contest with heterogeneous risk-averse workers, multiple prizes should be rewarded to maximize the requesters utility. In contrast \cite{heterogeneouscrowdsourcing} and \cite{crowdsourcingcontests} demonstrated that in an asymmetric all-pay auction-based contest of heterogeneous workers, requesters can maximize the contribution by rewarding only the top workers \cite{heterogeneouscrowdsourcing}. However, a winner-takes-all reward distribution may discourage risk-averse workers.\cite{tullock} mitigated this problem by introducing a lottery mechanism to give every player a strictly positive chance of winning as long as they participate. \cite{endogenous} analyzed equilibrium and optimal reward distribution for online Q\&A forums and competition platforms. \cite{payment} proposed an optimal reward policy (base salary + bonus) to optimize requesters profit where workers are selected based on workload demands and past performances. \cite{Truthfulincentives} proposed an optimal price setting for crowdsourcing by minimizing regret, varified on a Human Intelligent Task on Amazon Mechanical turk. \cite{Competitivegame} conducted large-scale real experiments to investigate how competitive and lottery reward policies affect the cost and time efficiency of crowdsourcing. \cite{simplecontest} conducted a series of experiments using contests to understand the effect of the workers’ strategy and determine whether they should participate in contests. 

\textbf{Applied Mechanism Designs in Federated Learning.} Other works investigated different mechanism designs in FL context to incentivize participation, namely correlated agreement \cite{CAinFL}, contract theory \cite{ContractMultiDim,FLContract,ContractReputationBlockchain}, stackelberg games \cite{StackelbergFL}, multidimensional auction games \cite{Multiauction1,FLMultiauction2} and Vickrey-Clarke-Groves design \cite{VCG}. \cite{DRL1, DLR2} used deep reinforcement learning (DLR) mechanism to find the optimal pricing strategy for the central server in context of information asymmetry. \cite{TemporalRegret} invented a novel FL incentive mechanism addressing the problem of temporary delay between contributions of workers and future rewards of an increased model, ensuring long-term fairness of distributing profit over time with multiple contributions. However, all of the aforementioned works assume a trustworthy central server/aggregator and do not allow for a decentralized architecture with equal power among peers.\newline

\textbf{Decentralized Federated Learning.}
Several works investigated Blockchain in FL as a tool to mitigate central coordination \cite{nature2021,Blockchained,DeepChain,9293091} while improving security and robustness \cite{DeepChain,GFL,9293091}.\cite{nature2021} introduced a decentral and confidential machine learning framework (swarm learning) on blockchain, showcasing that collaboration leads to better results on confidential medical data. \cite{FLBlockchainCrowdsourcing} introduced an incentivized crowdsourcing protocol atop public blockchains for machine learning tasks, where a purchaser can buy NN parameters of a trained model from workers. A mechanism design prevents workers from acting maliciously by being evaluated against coworkers. The implementation design and the theoretical analysis of the proposed MD lack detail. \cite{FLBlockchainMAIN} introduced a blockchain enabled FL framework incentivizing workers while utilizing contest theory to determine the optimal reward distribution. Applying \texttt{FedAvg}, every worker has to peer review models of every other worker in order to send his vote of the best workers to the smart contract, causing overhead on the workers' side while preventing the model from scaling. \cite{DeepChain} invented a novel blockchain system for Federated Learning, providing data confidentiality and auditability. During the \texttt{FedAvg} training process, all gradients are encrypted and stored on the permissioned blockchain, causing computational and storage overhead. How contributions are evaluated for the reward mechanism is not explicitly specified. \cite{ContractReputationBlockchain} introduce a novel mechanism, combining contract theory and reputation to incentivize high performance workers to participate. The blockchain serves as auxiliary entity storing the reputation of each worker. \cite{FLChain} introduces a novel decentralized FL framework (FLChain) to replace the parameter server, where workers get rewarded. How workers contributions are evaluated is not specified.
\cite{Liu2020} invented a new blockchain consensus system, where \texttt{FedAvg} participation is rewarded based on the Shapley Value (marginal contribution) by miners. \cite{Biscotti} proposed a blockchain-based privacy preserving ML platform, where model updates are stored on the blockchain, applying Shamir secrets for a secure gradient aggregation and differential privacy on the gradients. Table~\ref{tab:BlockchainFL} outlines the differences regarding the Federated Learning algorithm, the applied MD/incentive mechanism, the type of blockchain, whether the framework allows for different NN architectures on the client side, whether the framework induces a single point of failure (SPF) and whether the additional Blockchain system causes overhead. \newline

\begin{table}[]
    \centering
    \caption{Blockchain based Federated Learning Systems with Incentives.}
    \begin{tabular}{c|c|c|c|c|c|c|c}
       & FL & Reward Mechanism & Blockchain & NN agnostic & SPF & Scalability \\
    \toprule
        \cite{nature2021}& \texttt{FedAvg} & N/A & ASBS & no & no &  limited \\
        \cite{FLBlockchainCrowdsourcing} & N/A & peer review &  agnostic & N/A & yes &  N/A \\
        \cite{FLBlockchainMAIN} & \texttt{FedAvg} & contest theory & agnostic & no & no & limited \\
        \cite{DeepChain} & \texttt{FedAvg} & value based/no MD &  ASBS & no & no & limited \\
        \cite{ContractReputationBlockchain} & \texttt{FedAvg} & contract theory/reputation  &  (BC not for FL) & no & yes & limited \\
        \cite{FLChain} & \texttt{FedAvg} & no MD & ASBS & no & no & limited \\
        \cite{Liu2020} & \texttt{FedAvg}  & Shapley value &  ASBS & no & no & limited \\
        \cite{Biscotti} & \texttt{FedAvg} & N/A &  ASBS & no & no & limited\\
        \cite{Blockchained} & \texttt{FedAvg} & N/A & ASBS & no & no & limited \\
        this work & FD & PTSC &  agnostic & yes & no & good
    \end{tabular}
    \label{tab:BlockchainFL}
\end{table}

\begin{table}[]
  \caption{Notation Table}
  \label{tab:freq}
  \begin{tabular}{ccl}
    \toprule
    Symbol&Definition\\
    \midrule
    $\mathcal{S}$ & Governing smart contract deployed blockchain\\
    $\mathcal{C}$ & Set of classes of the public dataset $X^{pub}$\\
    $\mathcal{W}$ & Set of registered workers who register and deposit to the smart contract $\mathcal{S}$\\
    $\mathcal{W'}$ & Subset of workers contributing in training the model.$\mathcal{W'}\subseteq\mathcal{W}$ \\
    $\mathcal{F}$ & Federation of all possible workers\\
    $D_i$ & Reward staked by worker $i$\\
    $\mathcal{D}$ & Total deposit by all workers on the smart contract $\mathcal{D} = \sum_i\mathcal{D}_i$\\
    $V_{i}$ & Reward paid to worker $i$  \\
    $\mathcal{V}$ & Total reward paid out: $\mathcal{V} = \sum_i V_{i}$\\
    $m$ & Size of the public dataset $m = |X^{pub}|$ \\
    $n$ & Number of workers contributing to the training process. $n = |\mathcal{W'}|$ \\
    $n^{peers}_{j}$ & Number of peer workers who classify the same sample $j$\\
    $x_{ij}$ & Class prediction of worker $i$ on sample $j$, $x\in \mathcal{C}$\\
    $labelCount_{i}$ & Vector of occurences of each class in $Y^{1bit}_{i}$ of worker $i$\\
    $R(x)$ & Discrete density function of class $x\in\mathcal{C}$ in $X^{pub}$, $R(x)=\frac{\operatorname{labelCount}(x)}{\sum_{y} \operatorname{labelCount}(y)}, \forall y \in \mathcal{C}$\\
    $R_{i}(x)$ & R(x) without worker $i$'s contribution \\
    $i$ & Worker $i \in \mathcal{W}$ \\
    $j$ & Sample $j$ of the public dataset $X^{p u b}$ \\
    $\alpha$ & Parameter of the Dirichlet distribution to control data distribution\\ 
    $\beta$ & Penalty term of the reward function \\
    $\lambda$ & Reward scaling parameter. We set $\lambda=1$ across all experiments\\
    $\tau_{ij}$ & Reward for worker $i$ for the classification of sample $j$\\ 
    $\bar{\tau_{i}}$ & Total reward for worker $i$ \\
    $\mathbb{A}_{ij}$ & Heuristic to approximate certainty of the evaluated label $j$\\
    $e_{i}$ & Exerted effort by worker $i$\\ 
    $c_{i}(e_{i})$ & Variable cost incurred by exerting effort to classify $X^{pub}$ \\
    $ c_{\mathcal{S}}^{f i x}$ & Implicit cost for having access to $Y^{pub}_{aggr}$  \\
    $X^{pub}$ & Public dataset\\
    $X^{priv}_{i}$ & Private dataset of worker $i$\\
    $Y^{pub}_{i}$ & Soft (float) predictions of worker $i$ on $X^{pub}$. $Y^{pub}_{i}$ = $\mathbb{R}^{n\times|\mathcal{C}|}_{+}$ \\
    $Y^{priv}_{i}$ & Labels of the private dataset of worker $i$\\
    $Y^{1bit}_{i}$ & Integer encoded 1 bit quantized predictions of worker $i$ on $X^{pub}$.  $Y^{1bit}_{i}$ = $\mathcal{C}^{n}$\\
    $Y^{pub}_{aggr}$ & Aggregated 1-bit labels on smart contract $\mathcal{S}$ by majority vote. $Y^{pub}_{aggr}$ = $\mathcal{C}^{n}$\\
    \midrule
    $\mathcal{H}(\cdot)$ & 256bit SHA-3 Hash Function\\
    $salt_{i}$ & A random number \\
  \bottomrule
\end{tabular}
\end{table}

\section{Problem Statement and Reward Mechanism}

\subsection{Problem Statement}
We assume a federation $\mathcal{F}$ of workers $\mathcal{W}$ who have a common interest in advancing their private Neural Networks based on (i) additional data of other participants and (ii) the unlabelled public dataset $X^{pub}$ through Federated Distillation (FD). We consider an environment where all participants of $\mathcal{F}$ have equal power, e.g. no central entity such as a central server should have the power to either censor or manipulate the reward distribution. Each worker participating in the training is responsible for submitting predictions on public dataset $X^{pub}$ based on locally trained model, and label distribution $labelCount_{i}$ of the predictions. To enable decentralization, a smart contract atop of a blockchain will replace the central server (i) to aggregate the worker's predictions and (ii) to calculate the rewards considering other contributions. To ensure accountability and to prevent free-riding, each worker has to stake a deposit $D_{i}$. $\mathcal{D} = \sum_{i \in \mathcal{F}}D_{i}$ will be used to pay $\bar{\tau}_{i}$ for each contribution at the end of the training process. Note that $\bar{\tau}_{i} \geq D_{i}$ in case worker $i$'s contributions are above average to $\mathcal{F}$ and $\bar{\tau}_{i} \leq D_{i}$ otherwise. Malicious behavior like (i) withholding after committing, (ii) committing a wrong label distribution $labelCount_{i}$ will result in slash of deposit and exclusion from $\mathcal{F}$. The worker selection process is beyond the scope of this work, reputation \cite{ContractReputationBlockchain, reputationBC} or required registrations may be feasible solutions.

\subsection{Reward Mechanism Motivation}
\label{Sec:IntroRewardMechanism}
As no entity is in possession of the true labels of $X^{pub}$ in the decentralized Federated Learning setting, workers' evaluations cannot be verified. This might encourage workers to report random data without actually classifying $X^{pub}$. This can be mitigated by rewarding peer consistency, e.g. the reward depends on its consistency with the label given by other workers. However, the best strategy in such schemes is for all workers to report the same answer without investing effort in finding the real label. The solution to these issues is to set up a mechanism, where the expected profit for each individual worker is maximized, if they put high effort into solving the task while acting truthful. \newline

In contrast to a server-worker relationship, our framework assumes multiple stakeholders with common interest in improving their respective model. The initially staked deposit $D$ which will be used to pay $\tau$ manifests this mutual interest. Yet, contributions may be of different quality to the overall federation. Low quality workers may even have a negative effect on the overall federation even if their intention is truthful. At the same time, some classes in $X^{pub}$ may be less common and therefore are more important to classify correctly. Hence, a mechanism is required to:
\begin{enumerate}
    \item incentivize only workers with the best abilities for the task
    \item incentivize these workers to invest their utmost effort in obtaining the most accurate answer 
    \item incentivize workers who are able to classify uncommon samples in $X^{pub}$ with higher rewards
\end{enumerate}

We introduce the Peer Truth Serum for Federated Distillation (PTSFD), adopting Peer Truth Serum for Crowdsourcing (PTSC) \cite{PTSC} for the Federated Distillation environment. PTSC combines the reward mechanism of \cite{DasguptaGosh} with the idea of Peer Truth Serum \cite{PTS, PTS2} to  ensure incentive compatibility over a non binary solution space for heterogeneous workers. Because payment is a secondary motivation for workers, PTSFD introduces a penalty term $\beta$.  Workers get rewarded for each sample according to

\begin{equation}
\label{Eq:Reward_Tau}
        \tau_{ij}\left(x_{ij}\right)=\lambda \cdot\left(\frac{1}{n_{\text {peers }}} \sum_{p} \tau_{0}\left(x_{ij}, x_{pj}\right)-\beta\right)
\end{equation}
where $\lambda$ is a scaling parameter to adjust the magnitude of payment,  $\beta$ scales the reward-accuracy ratio and total reward $\bar{\tau}_{i} = \sum_{j\in X^{pub}}\tau_{ij}$ and 
\begin{equation}
\tau_{0}\left(x_{ij}, x_{pj}\right)=\left\{\begin{array}{ll}
\frac{1}{R_{i}\left(x_{ij}\right)} & \text { if } x_{ij}=x_{pj} \\
0 & \text { if } x_{ij} \neq x_{pj}
\end{array}\right.
\end{equation}

$R_{i}(x):\mathcal{C}\mapsto[0,1], \sum_{x\in\mathcal{C}}R_{i}(x)=1 $ represents the discrete density function, excluding worker $i$'s contribution, where $R_{i}(x)$ denotes the fraction of reported labels $R_{i}(x)=\frac{\operatorname{labelCount}(x)}{\sum_{y\in \mathcal{C}} \operatorname{labelCount}(y)}$.
\newline

\subsection{Gametheoretic Analysis}
\label{SubSec:Gametheoretic Analysis}

\textbf{Task.} We consider a crowdsourcing scenario in which a group of workers solves n statistically independent tasks, where a task refers to classifying every sample $j$ for all $j \in X^{p u b}$. The setting can be considered a two staged game. In stage 1, workers choose the amount of effort $e$ they want to invest in classifying $X^{p u b}$, e.g. the complexity of the NN, the amount of data, the number of training rounds, etc. In stage 2, workers decide on what to report. To simplify the analysis, we assume two levels of effort high $e_{1}$ and low $e_{0}$, where $e_{1}$ is the best work possible exerted by an honest worker and $e_{0}$ can intuitively be seen as a random answer without any effort put into it. The baseline model assumes each worker solves every task. Without loss of generality, workers can be randomly allocated to solve tasks s.t. each sample of $X^{p u b}$ is classified by at least two different workers. \newline

\textbf{Workers.} We assume workers to be individually rational, aiming to maximize expected profit $\Pi_{i} = Rewards_{i} - Costs_{i}$:
\begin{equation}
    \max \mathbb{E}\left(\Pi_{i}\right) = U_{i}(\theta_i^{improved}) + U_{i}\left(\bar{\tau_{i}} - [c_{i}(e_{i}) + c_{\mathcal{S}}^{f i x}]\right)
\end{equation}

$U_{i}$ represents the expected utility function of worker $i$, which can be different for every worker. The expected rewards of contributing to the federation $\mathcal{F}$ are twofold: (i) the expected utility of the own improved model $\theta_i^{improved}$ and (ii) the utility of the expected monetary reward from $\mathcal{S}$ for contributing to classify $X^{p u b}$. The training process causes variable costs $c_{i}(e_{i})$, where $c_{i}$ is an increasing function of effort $e_{i}$, that is $c_{i}(e_{i,1}) > c_{i}(e_{i,0})$, where $e_{i,0}$ denotes no effort and $e_{i,1}$ denotes high effort of worker $i$. Without loss of generality, effort represents the quality and quantity of private data, the quality of the model, number of training iterations, etc. In addition to the variable costs of actively contributing to the federation, fixed participation costs $c_{\mathcal{S}}^{f i x}$ are required to offset free-riding of inactive but registered workers of $\mathcal{S}$ who reap the benefit of an improved model $U_{i}(\theta_i^{improved})$ without contributing to the benefit of $\mathcal{F}$. Note that the initially staked Deposit will be used to pay for contributing workers, therefore $c^{fix}_{\mathcal{S}} = D_{i}^{before}-D^{after}_{i}$ describes the implicit costs for having access to $Y^{pub}_{aggr}$. 
\newline

\textbf{Incentive Compatibility.} In order to evaluate PTSFD in game theoretic terms, we analyze each workers expected profit $\Pi_{i} = Rewards_{i} - Costs_{i}$. We assume individual rationality (IR), e.g. workers try to maximize their expected profit and do not participate if $\Pi \leq 0$. For the sake of simplicity, we further assume that the gain in model improvement $U_{i}(\theta_{i}^{improved})$ is offset by $U_{i}(c^{fix}_{\mathcal{C}})$. \
When a worker classifies a sample, it obtains an evaluation $Y_{j}^{eval}$ which can be different from the reported value $Y_{j}^{report}$. In stage two, workers face three different strategies $\forall j \in X^{p u b}$ \cite{PTSC}:

\begin{enumerate}
    \item \textbf{Honest} Invest high effort $e_{1}$ to obtain $Y_{j}^{eval}$ and report honestly, s.t. $Y_{j}^{report} = Y_{j}^{eval}$
    \item \textbf{Strategic} Invest high effort $e_{1}$ to obtain $Y_{j}^{eval}$ but reports $Y_{j}^{report} \neq Y_{j}^{eval}$
    \item \textbf{Heuristic} Do not invest any effort $e_{0}$ and randomly report $Y_{j}^{report}$ based on the a-priori known distribution of labels in $X^{p u b}$ 
\end{enumerate}

We define the mechanism to be incentive compatible, if the honest strategy is the dominant strategy for every worker. We use an equilibrium analysis to determine the resulting behavior of each worker. In particularly, $\sigma = \left(\sigma_{1}, \sigma_{2}, \dotsc, \sigma_{n}\right)$ represents a strategy profile of each worker. This profile is an equilibrium $\bar{\sigma}$ if for any worker $i \in \mathcal{W}$, the workers expected profit is maximized with the honest strategy profile $\bar{\sigma}$. Suppose that worker $i$ believes that the peer workers are honest and their answer on a given sample $j$ is positively correlated with the worker $i$'s answer x, when obtained with high effort $e_{1}$. Specifically, worker $i$ believes that answer x is not less likely for sample $j$ than in the distribution over all tasks. \newline

\textbf{Honest Strategy.} For every sample $j$ in $X^{p u b}$, the worker calculates the probability scores over all possible classes in $\mathcal{C}$ (output of the softmax layer of a NN). Let us further assume worker $i$ is in possession of a trained model $\theta_{i}$, with an overall accuracy $Accuracy_{\theta_{i}}$. We define the relative certainty $\mathbb{A}_{ij}$ of any prediction of client i on an element $j$ of $X^{pub}$ as the product of the local classifier accuracy and the sample-specific maxprobabilityscore. 
\begin{equation}
\label{eq:Aj}
    \mathbb{A}_{ij} = Accuracy_{\theta_{i}} \cdot MaxProbabilityScore_{ij} 
\end{equation}

Under the assumption that the local client data $X^{priv}_{i}$ is representative of the entire data distribution D, this metric will give a heuristic measure for the data specific certainty in the model prediction. Based on this metric, each worker will make the decision whether to report predicted labels, discarding those for which reward is expected to be negative. This leads to the expected profit
\begin{equation}
\label{Eq:Expected Reward}
    \mathbb{E}(\Pi_{ij}) = \mathbb{A}_{ij}\cdot \lambda \left(\frac{1}{R(x_{ij})}-\beta\right)+(1-\mathbb{A}_{ij}) \cdot \lambda (-\beta) - c_{i}(e_{i})
\end{equation}

Assuming individual rationality, $\mathbb{E}(\Pi_{i,j}) \geq 0$ in order to incentive worker $i$ to submit a vote on sample $j$. Following ~\ref{Eq:Expected Reward}, we can derive minimum prediction quality 
\begin{equation}
\label{Eq:MinimumAj}
     \mathbb{A}_{ij} \geq R(x_{ij})\cdot\left(\frac{c_{i}(e_{i})}{\lambda}+\beta\right)
\end{equation}
 required to incentivize worker $i$ to participate, e.g. $\Pi_{i} \geq 0$. Notice that the federation can set the overall quality threshold by adjusting hyperparameter $\lambda$ and $\beta$ appropriately, assuming similar variable costs $c(e)$ on the workers side.\newline 

\textbf{Heuristic Strategy.} The heuristic strategy assumes worker $i$ does not put in any effort to obtain $Y_{j}^{eval}=x \in \mathcal{C}$. The expected reward depends on the probability of matching the peer's answer, where the answer x is independent of the task. Thus, the probability of matching a peer coincidentally is equal to the frequency of an answer $x \in \mathcal{C}$. 
\begin{equation}
    \mathbb{E}(\Pi_{ij}) = R(x_{ij}) \cdot\left(\frac{1}{R(x_{ij})}-\beta\right)+(1-R(x_{ij})) \cdot(-\beta)=1-\beta
\end{equation} 
Note that the expected profit for $\beta=1$ is 0 and strictly negative for $\beta > 1$, independent from what the answer x is, or what the worker knows about the distribution $R(x)$ for each label over $X^{p u b}$. Since the noise added to classifying $X^{p u b}$ will lower overall model quality, following Equation~\ref{Eq:Expected Reward}, we can expect that a rational worker will not elect to participate in case $\beta\geq1$.\newline  

\textbf{Strategic Strategy.} Under the assumption of honest participation of other workers, exerting $Y_{j}^{eval}$ while reporting $Y_{j}^{report} \neq Y_{j}^{eval} $ will result in a negative expected profit $\forall j \in X^{p u b}$ 
\begin{equation}
    \mathbb{E}(\Pi_{ij}) = \lambda \cdot(-\beta) - c_{i}(e_{i})
\end{equation} 

as long as the self-predicting condition \cite{PTSC} holds, e.g.
\begin{equation}
\label{eq:SelfPredictingCondition}
\frac{A_{ij}(x|x)}{R(x)} > \frac{A_{ij}(\bar{x}|x)}{R(\bar{x})}, \forall \bar{x} \neq x 
\end{equation}

Let us consider the case when workers collude, that is, they report x when $Y_{j}^{eval} = x$ and when $Y_{j}^{eval} = y$. However, R will change accordingly, that is $R^{col}(x) = R(x)+ R(y)$

\begin{equation}
    \mathbb{E}(\Pi_{ij})=\left\{\begin{array}{ll}
\mathbb{A}_{ij}\cdot \lambda \left(\frac{1}{R_{i}(x_{ij})+R_{i}(y_{ij})}-\beta\right)+(1-\mathbb{A}_{ij}) \cdot \lambda (-\beta) - c_{i}(e_{i}) & \text { if } Y_{ij}^{eval}=x_{ij} \\
\mathbb{A}_{ij}\cdot \lambda \left(\frac{1}{R_{i}(x_{ij})+R_{i}(y_{ij})}-\beta\right)+(1-\mathbb{A}_{ij}) \cdot \lambda (-\beta) - c_{i}(e_{i}) & \text { if } Y_{ij}^{eval}=y_{ij} 
\end{array}\right.
\end{equation}

and exactly offset the gain in matching probability. Therefore, only honest strategy with a high quality model will lead to a positive expected reward for the respective worker. This leads to an equilibrium $\bar{\sigma}^{honest}$ of the PTSFD mechanism, which proves incentive compatibility.

\section{1-bit Compressed Federated Distillation Framework with Smart Contract Logic}
The protocol contains the following steps: (i) Task Specification \& Contract Deployment, (ii) Worker Registration \& Deposit, (iii) Local Model Training, (iv) Hash Commit Submission, (v) Reveal Predictions, (vi) Aggregation \& Reward Distribution, and (vii) Knowledge Distillation from $X^{pub}$.

\begin{figure}[H]
    \centering
    \includegraphics[scale=0.3]{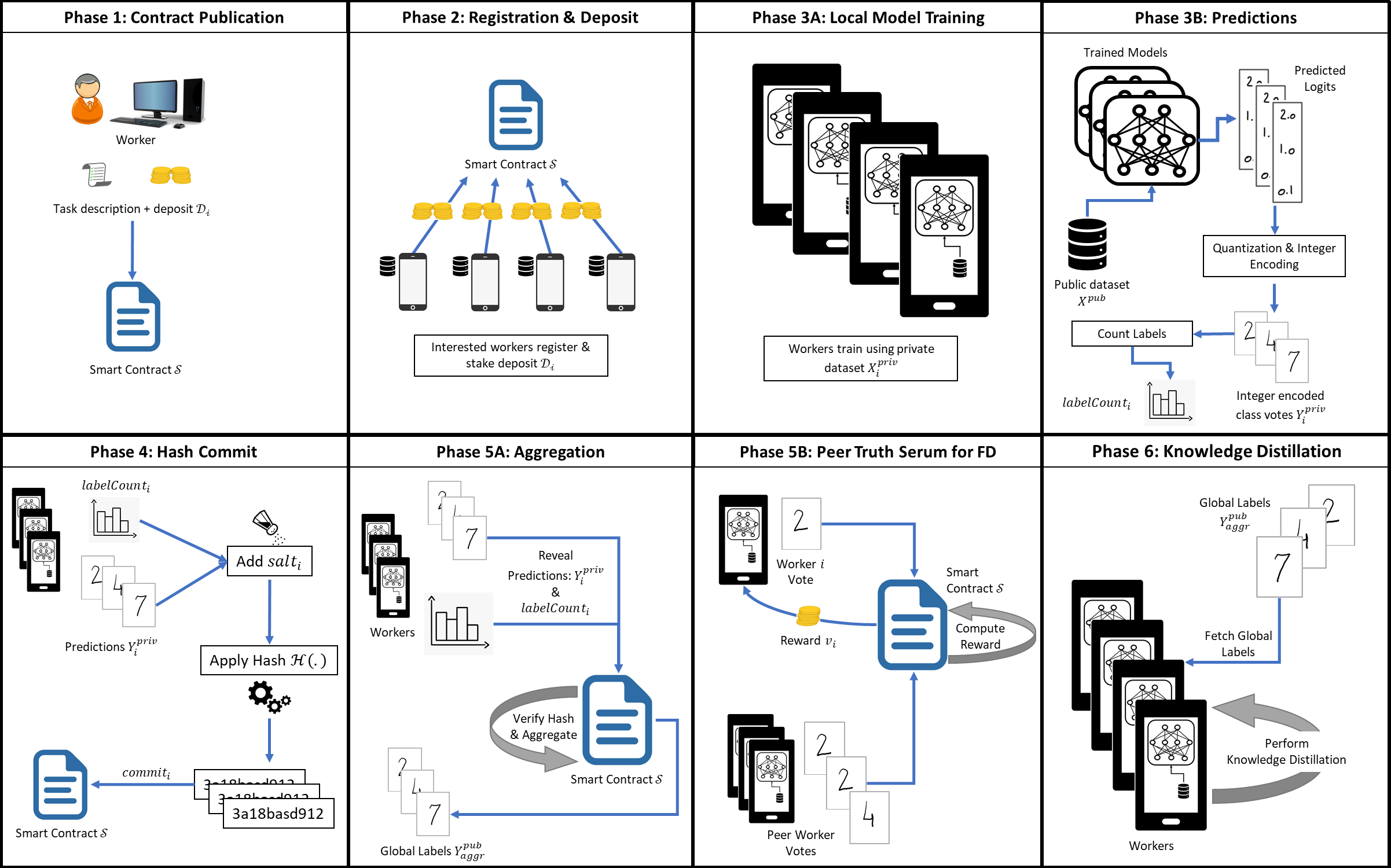}
    \caption{Iterative Process of on-blockchain Federated Distillation.}
    \label{fig:my_label}
\end{figure}

\subsection{Task Specification \& Smart Contract Deployment}
In order to form Federation $\mathcal{F}$, participants with similar interests need to agree upon the requirements and specifics of a FD task, namely:
\begin{enumerate}
    \item Task description and data distribution (e.g. images of a certain type).
    \item Reference to a public data set $X^{p u b}$ and possible classes $\mathcal{C}$ for the Federated Distillation pipeline, which will later be used by the workers to predict the labels on each sample of the dataset. 
    \item Reference to the address of $\mathcal{S}$.
    \item Deposit amount $D_{i}$ which has to be staked by every worker.
    \item PTSFD and reward mechanism details ($\lambda$ and $\beta$ values).
\end{enumerate}

Once a federation $\mathcal{F}$ is formed, either an external third party or any of the workers of $\mathcal{F}$ deploys the governing smart contract $\mathcal{S}$ and stakes the required deposit $D_{i}$, the addresses of all viable workers $\in \mathcal{F}$ as well as the aggregation and PTSFD logic of the FD task. 

\subsection{Worker Registration \& Deposit Submission} Based on the task specifications, interested workers register on the smart contract $\mathcal{S}$ with their respective blockchain address (public key) and match the required deposit $D_{i}$. $\mathcal{S}$ checks whether the applying worker is part of the federation. Assuming $|\mathcal{F}| >> |\mathcal{W}|$, PTSFD encourages workers of high value for $\mathcal{F}$ in terms of data and computational capacity to participate while discouraging low quality workers as we will show in Section~\ref{Sec:Reward Mechanisms}. To prevent free-riding from workers in $\mathcal{F}$ who are not registered to participate should not have access to $\mathcal{S}$. This can be achieved by deploying $\mathcal{S}$ on an appropriate blockchain system or through shuffling of $X^{pub}$ s.t. only participating clients have access to the correct indices. 

\subsection{Local Model Training and Prediction} The total training process contains two phases, local model training phase on local data $X^{priv}_{i},Y^{priv}_{i}$ and the Knowledge Distillation phase from $X^{pub}, Y^{pub}_{aggr}$ as the last step of the protocol, as outlined in Section~\ref{Sec:Federated Distillation}. \newline
\textbf{Training on Local Data.} Each worker is either in possession of a pre-trained model or starts training a NN locally on their respective private data until convergence (optional: until an initially agreed minimum accuracy among $\mathcal{F}$). Note that in contrast to \texttt{FedAvg}, FD does not require the same shared NN architecture among all workers, which allows workers to choose an optimal architecture with respect to their computational resources.\newline
\textbf{Label Prediction.} After the the training process, workers will then calculate the soft labels $Y_i^{pub} = \{f_{\theta_i + \Delta\theta_i}(x)\ |\ x \in X^{pub}\}$  
and then quantize these to 1bit ${Y}_i^{1bit} = Q_{1bit}(Y^{pub}_i)$. \newline 
\textbf{Label Count.} Because the PTSFD mechanism requires information about the label distribution $R(x)$ over $X^{p u b}$ to calculate rewards, each worker $i$ is required to calculate the label count $labelCount_{i} \in \mathbb{N}^{|\mathcal{C}|}$ of each label found in $X ^ {p u b}$, to mitigate computational overhead on blockchain (outlined in Algorithm~\ref{Alg:Label Count Report}). The additional validation function to check the correct calculation of $labelCount_{i}$ depends on the underlying blockchain system and is beyond the scope of this work.  \newline

\IncMargin{1em}
\begin{algorithm}[H]
\DontPrintSemicolon

\SetKwData{Left}{left}\SetKwData{This}{this}\SetKwData{Up}{up}
\SetKwProg{Init}{init}{}{}
\SetKwInOut{Input}{input}
\SetKwInOut{Output}{output}
\Input{Integer encoded class votes $x_{ij}$, where $i \in \mathcal{W'} \subseteq \mathcal{W}$}
\Output{$labelCount_{i}$}
\BlankLine
\Init{$labelCount_{i}$}{var \texttt{$labelCount_{i}$} $\in \mathbb{N}^{|\mathcal{C}|} = \left(0, 0, \ldots, 0\right)$\;}
\BlankLine
\ForEach(\tcp*[f]{iterate over data samples}){$j \in X^{pub}$} {
$labelCount_{i}(x_{ij}) \pluseq 1$\;
}

\Return{$labelCount_{i}$}\;

\caption{Local label count for worker $i$}
\label{Alg:Label Count Report}
\end{algorithm}\DecMargin{1em}

\subsection{Commit and Reveal} Information on blockchain is transparent to every node. Even in a private blockchain setup, workers in $\mathcal{W}$ could wait for peers to publish $Y^{1bit}_{p}$ and copy their results without putting in any effort. To prevent copying and to force workers to exert effort to classify $X^{pub}$, we apply a two-step commit and reveal scheme. 

\begin{enumerate}
    \item \textbf{Commit.} Before publishing the results to $\mathcal{S}$ where all peer workers would be able to see the submission, a cryptographic hash $hashCommit_{i}=\mathcal{H}\left(Y^{1bit}_{i}, salt_{i}, labelCount_{i}\right)$ is calculated to obfuscate $Y^{1bit}_{i}$ and $labelCount_{i}$. The property of pre-image resistancy of a cryptographic hash function (e.g. it should be difficult to find any message m such that $commit_{i} = \mathcal{H}\left(m\right)$) as well as the property of collision resistance (e.g. it should be difficult to find two different messages $m1$ and $m2$ such that $\mathcal{H}\left(m_{1}\right) = \mathcal{H}\left(m_{2}\right)$) ensures that no worker can either recover $Y^{1bit}_{i}$ nor later change their previously committed $Y^{1bit}_{i}$. Each worker $i$ sends $hashCommit_{i}$ to $\mathcal{S}$ as soon as it finishes training. Note that the commit phase on $\mathcal{S}$ ends once $|\mathcal{W'}| \subseteq |\mathcal{W}|$ workers have registered on $\mathcal{S}$ or the maximum time $T^{m a x}_{commit}$ has elapsed.
    \item \textbf{Reveal.} The reveal phase on $\mathcal{S}$ requires each worker who successfully committed in the commit-phase to reveal $Y^{1bit}_{i}$, $labelCount_{i}$ and $salt_{i}$ within time $T^{m a x}_{reveal}$ through a transaction function call on $\mathcal{S}$. To prevent withholding attacks, a worker deposit $D_{i}$ gets slashed if worker $i$ does not reveal within a sufficiently large time $T^{m a x}_{reveal}$. The smart contract checks whether the commit is viable, s.t. $\mathcal{H}\left(Y_{i}, labelCount_{i}, salt_{i}\right) == hashCommit_{i}$. 
\end{enumerate}

Algorithm~\ref{Alg:Commit Reveal} outlines the pseudo code of such a scheme in Solidity on the Ethereum blockchain.

\IncMargin{1em}
\begin{algorithm}[]
\DontPrintSemicolon
\KwData{$\mathtt{32byte}$ $hashCommit_{i}$ $\gets$ $\mathcal{H}\left(Y^{1bit}_{i},labelCount_{i},salt_{i}\right)$}
\SetKwProg{Init}{Init}{}{}
\SetKwProg{Mycommit}{Phase I}{}{}
\SetKwProg{Myreveal}{Phase II}{}{}
\SetKwFunction{Require}{require}
\SetKwFunction{Reveal}{reveal}
\SetKwFunction{Commit}{commit}
\SetKwFunction{reward}{keccak256}
\Init{}{
var \texttt{commitments} $\gets$ Mapping($address_{i}$ => byte32) $\forall i \in \mathcal{W}$ \;
var \texttt{userIsCommitted} $\gets$ Mapping($address_{i}$ => bool) $\forall i \in \mathcal{W}$ \;
}
\Mycommit{\Commit{hashCommit}}{
\ForEach{$i \in \mathcal{W'}$}{
\Require{userIsRegistered(msg.sender)}\tcp*[f]{registered in $\mathcal{W}$}\;
\Require{!userIsCommited(msg.sender)}\;
    $commitments.append(commit_{i})$\;
    $isCommitted(msg.sender) = True$\;
    }
}

\Myreveal{\Reveal{$Y^{1bit}_i, salt$}}{
\ForEach{$i \in \mathcal{W'}$}{
\Require{userIsCommitted(msg.sender)}{}\;
\Require{$\mathcal{H}\left(Y^{1bit}_i, labelCount_{i}, salt_{i}\right) == commitments(msg.sender)$}{}\;
    }
}

\caption{Commit and Reveal Protocol}
\label{Alg:Commit Reveal}
\end{algorithm}
\DecMargin{1em}

\subsection{Aggregation \& Reward Distribution} 
\label{Sec:Reward Mechanisms}
We apply PTSFD to calculate the reward distribution for each worker. In order to calculate the rewards, $\mathcal{S}$ aggregates $labelCount_{i}$ across all workers $i\in\mathcal{W'}$ first to obtain the global label count $G = \sum_{\mathcal{W}'}{labelCount_i} \in\mathbb{N}^{|\mathcal{C}|}$ . G is a helper variable to calculate $R_{i}$:
\begingroup\abovedisplayskip=10pt \belowdisplayskip=10pt
\begin{equation}
R_i = \displaystyle \frac{1}{m \times n}\times\left(G - labelCount_{i}\right)
\end{equation}
\endgroup
The worker is rewarded for its prediction on sample $j$ with respect to it's peers regarding Equation ~\ref{Eq:Reward_Tau}. The final \textit{rewardScore} for worker $i$ is a sum of all individual rewards over $X^{pub}$, given by

\begingroup\abovedisplayskip=10pt \belowdisplayskip=10pt
\begin{equation}\label{Eq:Reward_Score}
\bar{\tau_{i}} = rewardScore(i) = \lambda \cdot \left( \frac{1}{n^{peers}_{j}} \sum_j \sum_p \tau_0(x_{ij},x_{pj})\right) \forall i \in \mathcal{W'}
\end{equation}
\endgroup
\newline
where parameter $\lambda$ describes a scaling parameter for the reward and $n^{peers}_{j}$ describes the number of peer workers who also submitted a label prediction on $j$.
The aggregated predictions $Y^{pub}_{aggr}$ are calculated by majority vote of $Y^{1bit}_i$ $\forall i\in\mathcal{W'}$. We merge the reward computation and aggregation into a single algorithm as outlined in Algorithm~\ref{Alg:PTSFD}. Note that implementation details may differ fundamentally depending on the underlying blockchain architecture.

\IncMargin{1em}
\begin{algorithm}[]
\DontPrintSemicolon

\SetKwData{Left}{left}\SetKwData{This}{this}\SetKwData{Up}{up}
\SetKwProg{Init}{init}{}{}
\SetKwInOut{Input}{input}
\SetKwInOut{Output}{output}
\Input{Integer encoded class votes $x_{ij}$, and $labelCount_{i}$ where $\forall i \in \mathcal{W'} \subseteq \mathcal{W}$, $\forall j \in X^{pub} $ }
\Output{$rewardScore$, $globalLabels$}
\BlankLine
\Init{$rewardScore$, $globalLabels$, $Votes$, $G$, $M$, $R_i$,  $\tau_0$}{
var \texttt{$rewardScore$} $\in \mathbb{R}^{n}$\tcp*[r]{reward share of each worker}\;
var \texttt{$globalLabels$} $\in \mathbb{N}^{m}= \begin{pmatrix}0, & 0, & \ldots & 0\end{pmatrix}$\tcp*[r]{final labels for $X^{pub}$}\;
var \texttt{$Votes$} $\in \mathbb{N}^{m\times|\mathcal{C}|}=\begin{pmatrix}0 & 0 & \cdots & 0\\ \vdots & \vdots & & \vdots\\0 & 0 & \cdots &\; 0\end{pmatrix}$\tcp*[r]{placeholder: aggregated votes}\;
var \texttt{$M$} $\in \mathbb{N}^{n}=\begin{pmatrix}0, & 0, & \ldots & 0\end{pmatrix}$\tcp*[r]{placeholder: \# of predictions}
var \texttt{$G$} $\in \mathbb{N}^{|\mathcal{C}|}=\begin{pmatrix}0, & 0, & \ldots & 0\end{pmatrix}$\tcp*[r]{placeholder: data distribution}
}
\ForEach(\tcp*[f]{iterate over selected workers}){$i \in \mathcal{W'}$}{
$G \pluseq labelCount_{i}$\;
\ForEach(\tcp*[f]{iterate over data samples}){$j \in X^{pub}$}{
\If(\tcp*[f]{count only if predicted}){$c_{ij} \neq \texttt{NULL}$} {
$Votes[j,x_{ij}] \pluseq 1$\;
$M[i] \pluseq 1$
}}}

\BlankLine
\ForEach(\tcp*[f]{iterate over data samples}){$j \in X^{pub}$}{
\ForEach(\tcp*[f]{iterate over selected workers}){$i \in \mathcal{W'}$}{
\If(\tcp*[f]{skip if no prediction }){$c_{ij} = \texttt{NULL}$} {
\textit{continue}\;
}
\BlankLine
$R_i = \displaystyle \frac{1}{n \cdot M[i]}\times\left(G - labelCount_{i}\right)$\;
$\tau_0 = 0$\;
$n_{peers} = 0$\;
\BlankLine
\ForEach(\tcp*[f]{iterate over peers}){$p \in \mathcal{W'} \neq i$}{
\If(\tcp*[f]{skip if no prediction}){$c_{pj} = \texttt{NULL}$} {
\textit{continue}\;
}
\BlankLine
$n_{peers} \pluseq 1$\;
\BlankLine
$\tau_0 \pluseq \begin{cases}
            \frac{1}{R_i[x_{ij}]} - \beta & \text{if } x_{ij} = x_{pj}\\
            -\beta & \text{otherwise}.
        \end{cases}$\tcp*[r]{reward only if a match}}
$rewardScore[i] \pluseq \displaystyle \lambda \cdot \left[\frac{1}{n_{peers}}\times\tau_0\right]$\tcp*[r]{reward worker $i$ for sample $j$}}
\BlankLine
\For(\tcp*[f]{compute majority for sample $j$}){$c=0; c \leq |\mathcal{C}|; c++$} {
\If{$Votes[j,c] > globalLabels[j]$} {
$globalLabels[j] = c$\;
}}}

\Return{$rewardScore$, $globalLabels$}

\caption{Peer Truth Serum for Federated Distillation (PTSFD)}
\label{Alg:PTSFD}
\end{algorithm}\DecMargin{1em}

\subsection{Knowledge Distillation on Public Dataset} Finally, workers download the aggregated predictions $Y^{pub}_{aggr}$ from the blockchain and perform several epochs of knowledge distillation using $X^{pub}, Y^{pub}_{aggr}$ to improve their respective model ($\theta^{improved}_i \to \theta_i + \Delta\theta_i$). Optionally, "Local Model Training" -> "Hash Commit \& Aggregation Phase" -> "Reward Distribution" -> "Federated Distillation" can be repeated until a specific threshold is achieved as specified in the Smart Contract $\mathcal{S}$. Note that $\lambda$ should decrease for every consecutive round since most evaluated labels will not change. The details of the FD training process of each client is shown in Section~\ref{Sec:Federated Distillation}.

\subsection{Complexity Analysis}
Since we are running this protocol on-blockchain, it is imperative that the required computational and storage costs are well understood. Hence, in this section we discuss the overhead in terms of the computation and storage cost that our proposed algorithm incurs. Note again that the actual implementation on a general purpose Blockchain system may differ, depending on the underlying virtual machine. Yet, our PTSFD implementation illustrated in Algorithm~\ref{Alg:PTSFD} serves as a reference to approximate the complexity. 

\subsubsection{Computational Complexity} In Algorithm~\ref{Alg:PTSFD}, we first compute global label distribution and count class votes across all workers, this is done in first section (\textit{line 7 - 12}) of the algorithm. The computation overhead is $\mathcal{O}(m \cdot n)$ where $m=|X^{pub}|$ and $n=|\mathcal{W'}|$. Next we go over each data sample in $X^{pub}$ and reward/penalize a worker based on its peers. We also compute aggregated class label for each sample in this part of the algorithm (\textit{line 13 - 29}). The process of computing reward for each worker based on its peers incurs a computational overhead as given by $\mathcal{O}\left(m\cdot\sum_{i=1}^{n} n_{peers}\right)$. The global label calculation incurs an additional cost of $\mathcal{O}(m\cdot|\mathcal{C}|)$. In the baseline case, since each worker works on all data samples of the public dataset making it a peer of every other worker, the overall computation cost is given by Equation~\ref{Eq:Compute Cost}.
\begingroup\abovedisplayskip=10pt \belowdisplayskip=15pt
\begin{equation}\label{Eq:Compute Cost}
\mathcal{O}(m \cdot (n^2 + |\mathcal{C}|))
\end{equation}
\endgroup
For more practical solutions we distribute samples of public dataset among workers in a way that each sample is classified by a maximum of two workers. The computational cost of this implementation of PTSFD would reduce the overhead as described in Equation~\ref{Eq:Compute Cost 1}.
\begingroup\abovedisplayskip=10pt \belowdisplayskip=15pt
\begin{equation}\label{Eq:Compute Cost 1}
\mathcal{O}(m \cdot (2n + |\mathcal{C}|))
\end{equation}
\endgroup

\subsubsection{Storage Complexity} There are two types of storage cost associated with the proposed algorithm. One is the permanent storage cost, the other is that of non-permanent memory variables. $Votes$, $M$, $S$, $R_i$, $\tau_0$ require memory storage as part of the computation incurring $\mathcal{O}(|\mathcal{C}| \cdot (m + 2) + n)$ of additional memory storage. Whether reported frequencies $labelCount_{i}$ or final reward share of each worker $rewardScore$ have to be stored permanently on the blockchain depends on the requirements of the underlying blockchain system. In the optimal case, only $globalLabels = Y^{pub}_{aggr}$ is stored permanently on the blockchain. Therefore, the minimum bits of data required for each round is illustrated by Equation~\ref{Eq:Storage Costs}, where $\eta$ describes the additional overhead due to encoding requirements.
\begin{equation}
\label{Eq:Storage Costs}
\mathrm{b}_{globalLabels}=  n \times \left|\mathcal{C}\right| \times 1bit  + \eta
\end{equation}


\subsection{Limitations}
Despite the advantages of the introduced decentral FD protocol, our framework is restricted by the following limitations. 
\begin{enumerate}
    \item \textbf{Public Dataset.} Even though Federated Distillation introduces many advantages like reduced information exchange and independent NN architectures, the FD training process requires access to a public dataset $X^{pub}$ which might not be available for some use-cases. While \cite{FLChallenges} have shown that highly dissimilar data distributions can be sufficient for FD, relying on $\mathbb{A}_{ij}$ as a heuristic for the evaluation certainty of a sample restricts the divergence of distributions between $X^{pub}$ and $X^{priv}$. The scoring method introduced by \cite{sattler2021fedaux} seems promising in this context.
    \item \textbf{Public Blockchains.} Despite the heavy reduction of computational and storage requirements, our framework is not suitable for contemporary public blockchain systems due to (i) high costs induced by the storage capacity of $Y^{pub}_{aggr}$ and computational overhead of PTSFD as well as (ii) transparency of $Y^{pub}_{aggr}$ for nodes which are not part of  $\mathcal{W}\in\mathcal{F}$ and therefore did not deposit. Both problems might be mitigated by future developments in the public blockchain domain. 
    \item \textbf{Self Predicting Condition.} PTSFD is incentive compatible and leads to an optimal solution if workers act honestly. Yet, if Equation~\ref{eq:SelfPredictingCondition} holds, then the mechanism is incentive compatible. If classes are equally distributed over $X^{pub}$ the conditions always hold true. \textbf{Example}: Let Pr(x=a)=0,8 and Pr(x=b)=0,2 but R(a)=0,9 and R(b)=0.1. Even though worker $Y^{eval}_{i}=a$, their expected reward would be higher if $Y^{report}_{i}=b$ since $\frac{0,8}{0,9}-\beta < \frac{0,2}{0,1}-\beta$.

\end{enumerate}

\section{Experiments}
In this section we empirically evaluate the PTSFD framework and analyze the reward distribution under different levels of effort as well as its robustness in the event of malicious behavior. In this analysis, we do not consider explicit variable costs $c_{i}(e)$ since these are hard to quantify in most realistic scenarios. We further set the reward scaling parameter $\lambda = 1$ for all experiments. We do not consider lagging workers, therefore $\mathcal{W'}=\mathcal{W}$ across all experiments. All experiments are based on a single round of the proposed protocol.\\
\newline Specifically, we experimentally validate the following properties of PTSFD:
\begin{enumerate}
    \item \textbf{Performance.} Choosing to participate in the federation should lead to a significant improvement in model accuracy for each worker.
    \item \textbf{Fairness.} The more effort a worker exerts in terms of training accuracy and amount of training data, the better the reward.
    \item \textbf{Robustness.} Malicious workers are rewarded substantially less, even under high collusion rates.
\end{enumerate}
\subsection{Data sets and models}
\begin{figure}[H]
    \centering
    \includegraphics[width=\textwidth]{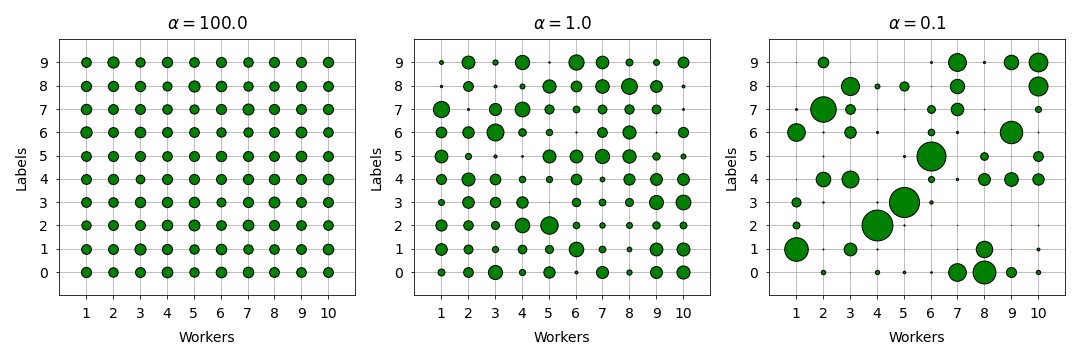}
    \caption{Non-iid data distribution case for different $\alpha$ values of Dirichlet Distribution.}
    \label{fig:Dirichlet Distribution}
\end{figure}

We analyze the decentralized 1-bit compressed FD with PTSFD protocol on a federated image classification problem, using EMNIST / MNIST data sets \cite{emnist}. Our Federation consists of 10 workers. We split the training data among workers according to a Dirichlet distribution with dirichlet parameter $\alpha$. Figure~\ref{fig:Dirichlet Distribution} illustrates the data distribution of 10 labels over 10 different workers for $\alpha=100$, $\alpha=1$ and $\alpha=0.1$.\newline We first train LeNet locally on $X^{priv}$ (which is EMNIST digits data set in our case) and then perform Knowledge distillation using the MNIST dataset as public $X^{pub}$ data set. Even though in real world PTSFD application, workers may train different model architectures and different local training epochs according to their own hardware constraints, we use only one default NN architecture for simplicity reasons and simulate heterogeneity through varying local training accuracy (early stopping), non-iid data and different sizes of $X^{pub}$. Note that the distribution of the distillation data deviates from the one of the worker data, as it would in realistic Federated Learning scenarios (MNIST contains handwritten digits, EMNIST contains different set of handwritten numbers). We use Adam optimizer \cite{adam} with a fixed learning rate of 0.001 for both the distillation and training process. We minimize cross-entropy loss for local model training on $X^{priv}, Y^{priv}$ and minimize Kullback-Leibler Divergence on $X^{pub}, Y^{pub}_{aggr}$. 

\subsection{Performance Improvement}
\begin{figure}[H]
    \centering
    \includegraphics[width=\linewidth]{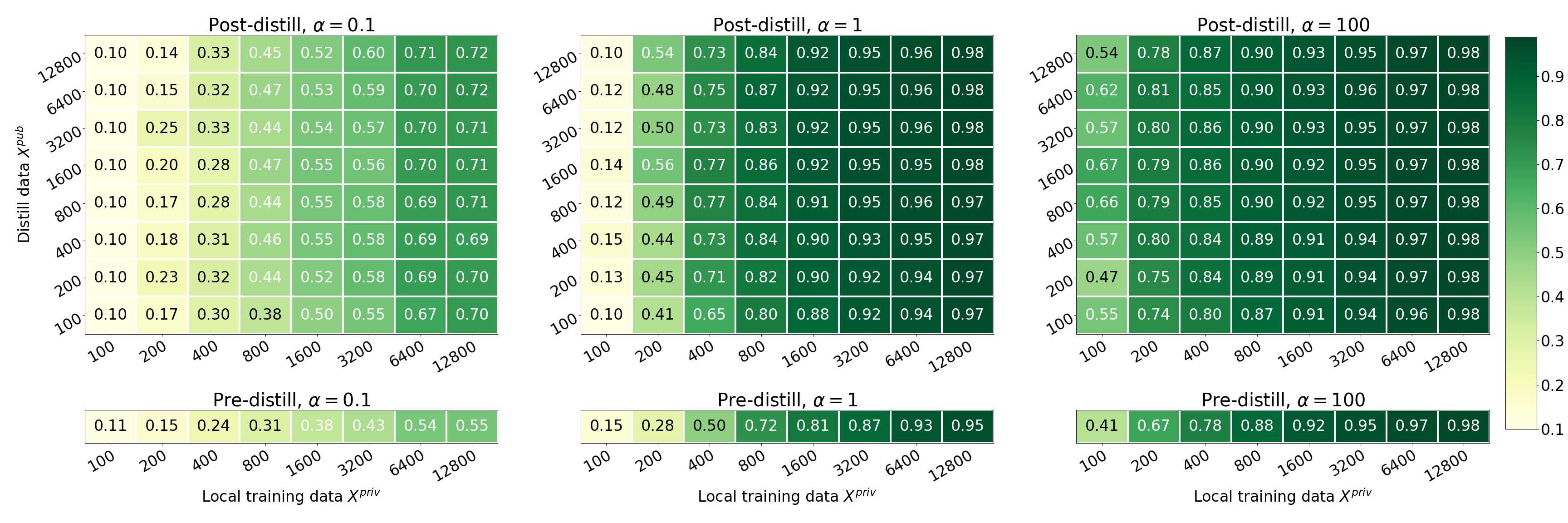}
    \caption{The influence of $X^{pub}$ and $X^{priv}$ on model accuracy under different $\alpha$ setting. }
    \label{fig:Heatmap1}
\end{figure}

In this section we evaluate participating workers for improvement in local model quality with varying the size of local training dataset and distillation dataset. We note a substantial increase in model quality for each worker after they run Knowledge distillation round. We also note that local training dataset size matters more than the distillation dataset size but cannot be ignored especially for non-iid distribution case.\newline
\newline Figure ~\ref{fig:Heatmap1} illustrates this increase in accuracy with respect to to the size of local dataset $|X^{priv}|$ and the public dataset $|X^{pub}|$ on EMNIST / MNIST with non-iid distribution ($\alpha = 0.1$ \& $\alpha = 1.0$) and with iid distribution ($\alpha = 100$).

\subsection{Fair Effort-Reward Correlation}
\begin{figure}[H]
    \centering
    \includegraphics[width=\textwidth]{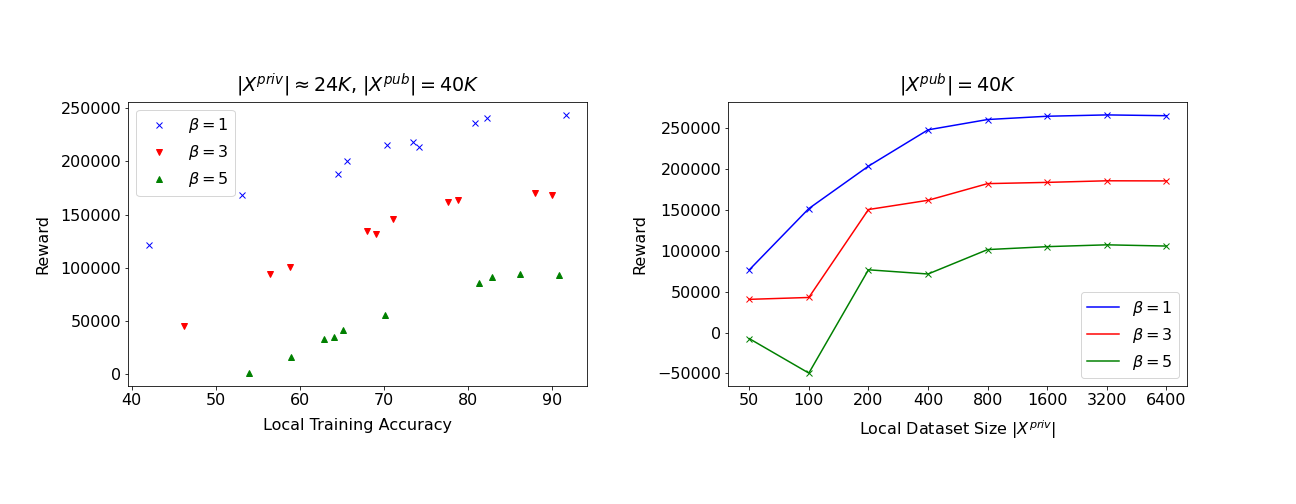}
    \caption{Effect of Local Training Accuracy \& Local Data Size on Reward.}
    \label{fig:RewardvsEffort}
\end{figure}

\textbf{Heterogeneous Effort.} For a realistic FL scenario, PTSFD allows for different levels of quality in terms of contributions. PTSFD workers can train different local model architectures with different number of training epochs according to their own hardware constraints. To mimic this heterogeneous behavior, we train 10 workers with different early stopping criteria. A high local training accuracy resembles high effort hence should yield better reward. We observe this in our experiments as shown by Figure ~\ref{fig:RewardvsEffort}.\newline
\newline\textbf{Heterogeneous Data Quantity.} The amount of private data as well as the respective quality of this data may vary and lead to different qualities of contribution. We assume similar data quality and assess different data quantities. 
Figure~\ref{fig:RewardvsEffort} illustrates reward distribution under heterogeneous data and accuracy measures. (left) shows the reward distribution of 10 workers with varying training accuracy under different $\beta$ values. (right) illustrates the effect of different amounts of local training data on the reward distribution. The results suggest that high training accuracy and a large local dataset lead to a higher reward.

\begin{figure}[]
    \centering
    \includegraphics[width=\textwidth]{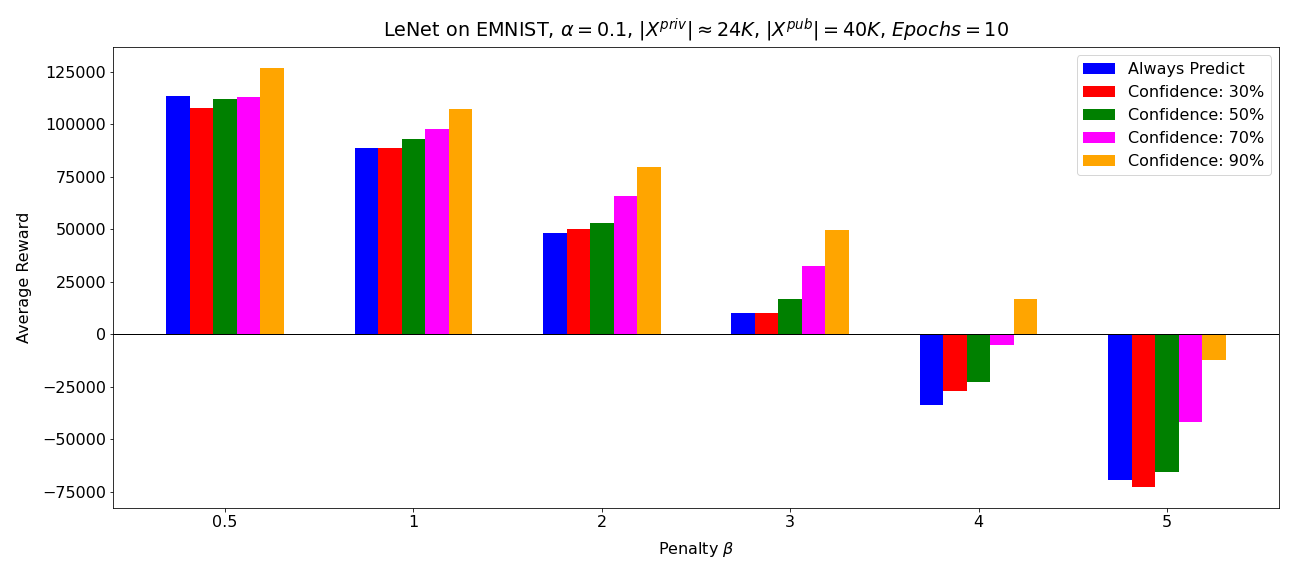}
    \caption{Effect of confidence based predictions on reward with different $\beta$ (penalty) using LeNet on MNIST.}
    \label{fig:confid_vs_reward}
\end{figure}
\subsection{Robustness of PTSFD}
In order to ensure a desired quality of label predictions, the federation can decide upon parameter $\lambda$ to scale the reward with respect to it's underlying collateral (we use $\lambda=1$ in all cases) and $\beta$ to tweak the penalty for wrong answers and therefore adjust the confidence necessary for individually rational workers (Equation~\ref{Eq:MinimumAj}) to submit a prediction. The initially staked deposit serves as safety mechanism against malicious behavior, since malicious behavior can result in negative gains. We design an experiment where each worker can choose to pass the report if they have low confidence in their predicted results. Fig \ref{fig:confid_vs_reward} shows the reward with different penalty factor $\beta$ under different confidence levels. In this experiment, we split the local training data according to a dirichlet distribution with dirichlet parameter $\alpha=0.1$ simulating a real world scenario where workers may not be in possession of homogeneous data. Therefore, a worker's local model may have a low ability to predict some classes previously not available to them. Workers will only report their result when their confidence on the most possible label exceeds a certain threshold. The results suggest that PTSFD can prevent low quality workers from polluting the federated training process by adjusting $\beta$ appropriately.

\subsection{Robustness in the case of Malicious Behavior}
\begin{figure}[H]
    \centering
    \includegraphics[width=\textwidth]{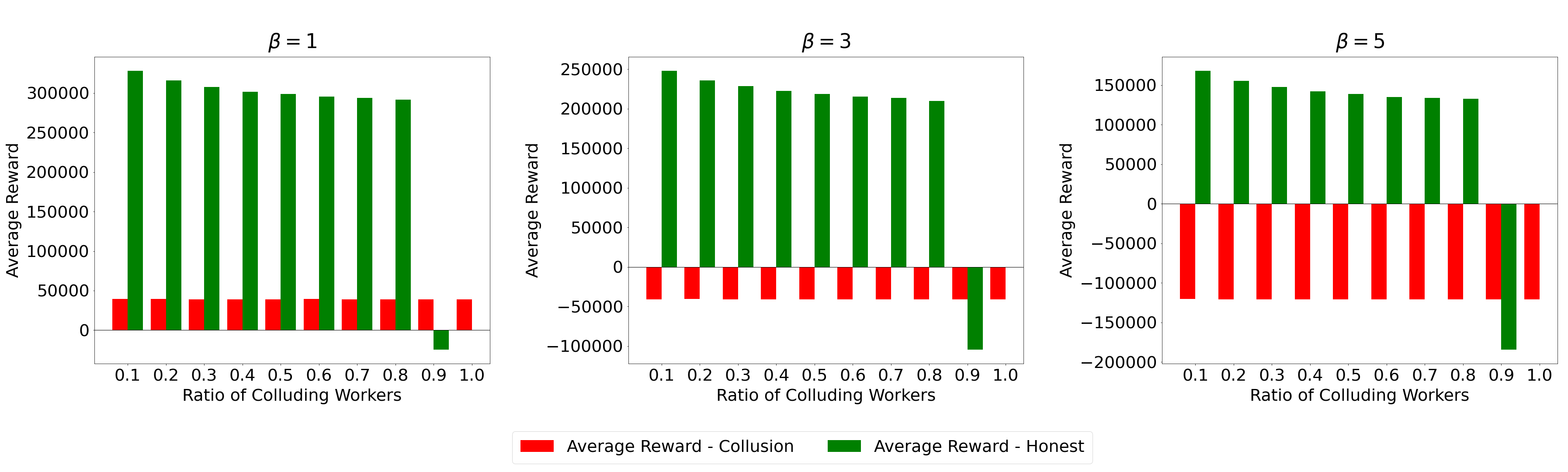}
    \caption{Average reward with varying ratio of colluding workers under different penalty $\beta$. Federated Learning setting with 10 workers running LeNet on EMNIST digits for 10 epochs. For all experiments, 40000 data points from the MNIST data set were used as public dataset $X^{pub}$.}
    \label{fig:Collusion}
\end{figure}

\begin{figure}[H]
    \centering
    \includegraphics[width=\textwidth]{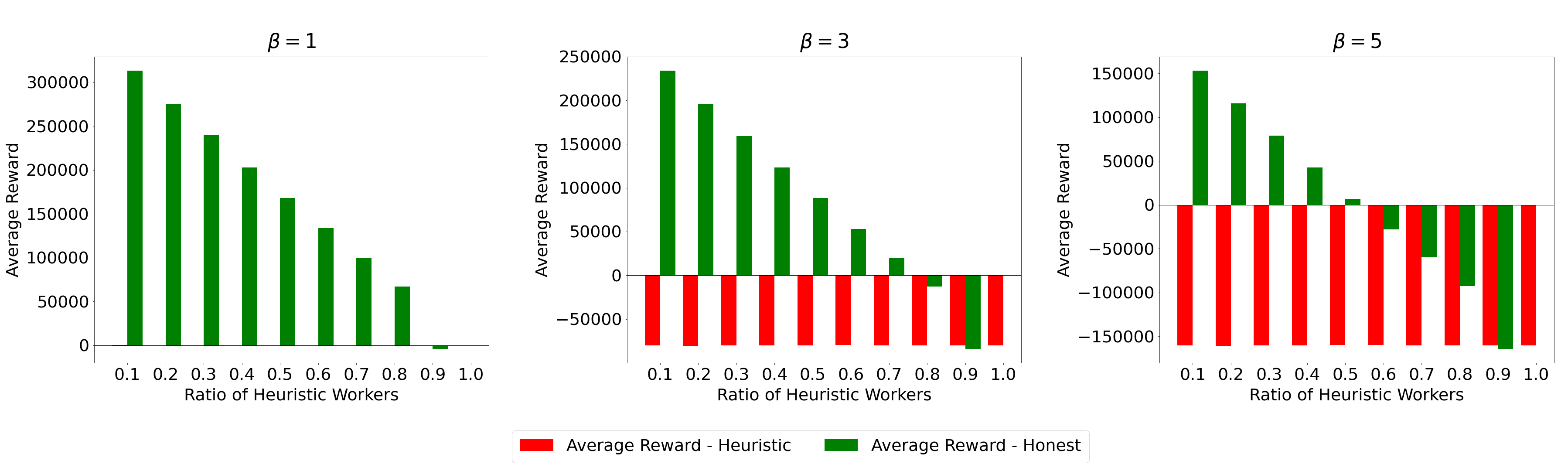}
    \caption{Average reward with varying ratio of heuristic workers under different penalty $\beta$. Federated Learning setting with 10 workers running LeNet on EMNIST digits for 10 epochs. For all experiments, 40000 data points from the MNIST data set were used as public dataset $X^{pub}$.}
    \label{fig:Heuristic}
\end{figure}

Finally, we experimentally verify the findings of the game-theoretic analysis presented in Section~\ref{SubSec:Gametheoretic Analysis} in a real FL context. We have theoretically proven that the heuristic behavior (skipping local training and reporting labels randomly on public dataset) as well as strategic behavior such as collusion results in an expected reward of $1-\beta$. The experiments verify our theoretical findings.\newline
\newline Figure \ref{fig:Heuristic} shows the average reward gained by heuristic workers versus the reward gained by honest participants. Heuristic workers predict on public dataset randomly instead of putting in any effort to train local model.\newline
\newline Figure \ref{fig:Collusion} illustrates the reward gains for colluding workers vs honest workers. Collusion is carried out by making predictions such that
$$
Y^{report}_{i}=\left\{\begin{array}{ll}
0 & \text { if } Y^{eval}_{i}\in \{0,1,2,3,4\} \\
9 & \text { if } Y^{eval}_{i}\in \{5,6,7,8,9\}
\end{array}\right.
$$
Both results suggest that honest participation yields the highest reward even if a large portion of workers act maliciously. An analysis where the additional costs are taken into consideration remains for future work.

\section{Conclusion}
In this work we have introduced a novel decentral and reward based 1-bit compressed Federated Distillation scheme on blockchain. We have shown, under various FL and non-iid conditions, that our proposed framework can lead to a substantial increase in model performance for every participant in the federation after only one round of participation. The 1-bit compression ensures explicit comparability between contributions, necessary to automatically compute rewards on a smart contract on top of a general purpose blockchain system in an environment where each worker is treated as an equal part of a federation. We have demonstrated that the reward distribution based on PTSFD, an adapted version of PTSC \cite{PTSC} is incentive compatible and enables the  federation to adjust to different thresholds of contribution quality by adjusting $\beta$. Both theoretical considerations and experimental evidence suggest that our proposed mechanism is robust against random reporting and collusion. We believe that our findings will help to scale Federated Learning tasks in fully decentralized environments where entities have an equal interest in improving their models.


\begin{acks}
This work was partly supported by the German Federal Ministry of Education and Research (BMBF) through the BIFOLD - Berlin Institute for the Foundations of Learning and Data (ref.\ 01IS18025A and ref.\ 01IS18037I) and the European Union’s Horizon 2020 Research and innovation Programme under Grant Agreement No.\ 957059.
\end{acks}

\bibliographystyle{ACM-Reference-Format}
\bibliography{sample-base}


\newpage
\setcounter{page}{1}
\appendix

\section{Game theoretic analysis}
The proposed framework requires each worker to stake a fixed amount $D_{i}$ to avoid malicious behavior. Unlike in crowdsourcing tasks where the reward can only be positive or zero, here any worker might be penalized with a negative reward $(\beta > 1)$. We follow the proof in \cite{2856102} and assume:
\begin{enumerate}
    \item \textbf{Self-predicting condition} \\ 
    \begin{center}
    $\frac{P_{p|w}(x|x)}{P_{p}(x)}>\frac{P_{p|w}(y|x)}{P_{p}(y)}, \forall y \neq x$
    \end{center}
    \item \textbf{Workers are rational agents aim to maximize expected reward} 
\end{enumerate}
In Section~\ref{Sec:IntroRewardMechanism} we introduced the PTSFD mechanism with the reward function: \\ 
$$
\tau_{ij}\left(x_{ij}, x_{pj}\right)=\lambda \cdot\left(\frac{1}{n_{\text {peers }}} \sum_{p} \tau_{0}\left(x_{ij}, x_{pj}\right)-\beta\right)
$$
$$
\tau_{0}\left(x_{ij}, x_{pj}\right)=\left\{\begin{array}{ll}
\frac{1}{R_{i}\left(x_{ij}\right)} & \text { if } x_{ij}=x_{pj} \\
0 & \text { if } x_{ij} \neq x_{pj}
\end{array}\right.
$$
Suppose that worker $w$ believes that the other workers are honest. The expected reward of worker $w$ for reporting $y$ while worker $w$ evaluates $x$ is equal to \\
\begin{center}
$\lambda(\frac{P_{p|w}(y|x)}{P_{p}(y)}-\beta)$
\end{center}
Where $P_{p|w}(y|x)$ represents the probability of peer $p$ reporting $y$ under the condition that worker $w$'s evaluation is $x$. $P_{p}(y)$ represents the probability of the peer $p$ reporting $y$.
Based on the initial self-predicting condition,  worker $w$ will always report $x$ when the other workers are honest, as long as\\ 
\begin{center}
$\lambda(\frac{P_{p|w}(x|x)}{P_{p}(x)}-\beta) > \lambda(\frac{P_{p|w}(y|x)}{P_{p}(y)}-\beta)$
\end{center}
Suppose that worker $w$ believes that the other workers adopt a strategy described by a distribution $Q_{p|p}$(strategic or heuristic strategies). 
$Q_{p|w}(y|x)$ represents the probability of peer $p$ reporting $y$ under the condition that worker $w$'s evaluation is $x$ \\
\begin{center}
$Q_{p|w}(y|x) = \sum_{z\subset{X}} Q_{p|p}(y|z)P_{p|w}(z|x)$ 
\end{center}
$Q_{p}(y)$ represents the probability of peer $p$ reporting $y$.
$Q_{p}(y)=\sum_{z\subset{X}}Q_{p|p}(y|z)P_{p}(z)$. The expected reward of worker $w$ for reporting $y$ while worker $w$ evaluate as $x$ is therefore equal to 
\begin{center}
$\lambda(\frac{Q_{p|w}(y|x)}{Q_{p}(y)}-\beta)$
\end{center}
We can rewrite the expected reward of worker $w$ whose evaluation is $x$ for reporting $y$ \\ 
\begin{center}
$\lambda(\frac{Q_{p|w}(y|x)}{Q_{p}(y)}-\beta)=\lambda(\frac{\sum_{z}Q_{p|p}(y|z).P_{p|w}(z|x)}{\sum_{z}Q_{p|p}(y|z).P_{p}(z)}-\beta)$
\end{center}
We can further expand the above equation into two terms.($z = x$ and $z \neq x$ )  \\

\begin{center}
$\lambda(\frac{Q_{p|p}(y|x).P_{p|w}(x|x)+\sum_{z \neq x}Q_{p|p}(y|z).P_{p|w}(z|x)}{Q_{p|p}(y|x).P_{p}(x)+\sum_{z \neq x}Q_{p|p}(y|z).P_{p}(z)}-\beta)\newline \newline
= \lambda(\frac{Q_{p|p}(y|x)*P_{p}(x).\frac{P_{p|w}(x|x)}{P_{p}(x)}+\sum_{z \neq x}Q_{p|p}(y|z).\frac{P_{p|w}(z|x)}{P_{p}(z)}.P_{p}(z)}
{Q_{p|p}(y|x).P_{p}(x)+\sum_{z \neq x}Q_{p|p}(y|z).P_{p}(z)}-\beta)\newline \newline
\leq \lambda(\frac{Q_{p|p}(y|x).P_{p}(x).\frac{P_{p|w}(x|x)}{P_{p}(x)}+\sum_{z \neq x}Q_{p|p}(y|z).\frac{P_{p|w}(x|x)}{P_{p}(x)}.P_{p}(z)}  
{Q_{p|p}(y|x).P_{p}(x)+\sum_{z \neq x}Q_{p|p}(y|z).P_{p}(z)}-\beta) \newline \newline
= \lambda(\frac{P_{p|w}(x|x)}{P_{p}(x)}-\beta)$
\end{center}

In the second line of the above equation we
multiply and divide $P_{p}(x)$ in the numerator. By applying the self-predicting condition on the second term in the numerator, we can get inequality in line three. We conclude that the maximum expected reward is obtained for honest reporting.





\end{document}